\documentclass{article}

\usepackage[preprint]{neurips_2024}

\usepackage[utf8]{inputenc} 
\usepackage[T1]{fontenc}    
\usepackage{hyperref}       
\usepackage{url}            
\usepackage{booktabs}       
\usepackage{amsfonts}       
\usepackage{nicefrac}       
\usepackage{microtype}      
\usepackage{xcolor}         
\usepackage{amsmath}
\usepackage{graphicx}
\usepackage{wrapfig}
\usepackage{amssymb}
\usepackage[utf8]{inputenc}
\usepackage{booktabs}  
\usepackage{graphicx}     
\usepackage{subcaption}   
\usepackage{caption}      

\newcommand{\red}[1]{\textcolor{red}{#1}}

\newcommand{\blue}[1]{\textcolor{blue}{#1}}

\newcommand{\violet}[1]{\textcolor{violet}{#1}}

\newcommand{\orange}[1]{\textcolor{orange}{#1}}
\definecolor{darkbrown}{rgb}{0.4, 0.26, 0.13}
\newcommand{\darkbrown}[1]{\textcolor{darkbrown}{#1}}
\newcommand{\boldsquare}[1][black]{%
  \textcolor{#1}{%
    \setlength{\unitlength}{1ex}%
    \begin{picture}(1.25,1.25)%
      \linethickness{0.5mm}%
      \put(0,0){\framebox(1.25,1.25){}}%
    \end{picture}%
  }%
}

\title{Understanding the Impact of Negative Prompts: When and How Do They Take Effect?}

\author{%
  Yuanhao Ban\\
  UCLA\\
  \texttt{banyh2000@gmail.com} \\
  \And
  Ruochen Wang \\
  UCLA\\
  \texttt{ruocwang@g.ucla.edu} \\
  \And
  Tianyi Zhou \\
  UMD \\
  \texttt{tianyi@umd.edu}
  \And
  Minhao Cheng \\
  PSU \\
  \texttt{mmc7149@psu.edu}
  \And
  Boqing Gong \\
  Google \\
  \texttt{bgong@google.com} \\
  \And
  Cho-Jui Hsieh \\
  UCLA \\
  \texttt{chohsieh@cs.ucla.edu} \\
}

\begin{document}

\maketitle
\vspace{-16pt}

\begin{abstract}
  The concept of negative prompts, emerging from conditional generation models like Stable Diffusion, allows users to specify what to exclude from the generated images.
Despite the widespread use of negative prompts, their intrinsic mechanisms remain largely unexplored. This paper presents the first comprehensive study to uncover how and when negative prompts take effect. Our extensive empirical analysis identifies two primary behaviors of negative prompts. \textit{Delayed Effect}: The impact of negative prompts is observed after positive prompts render corresponding content. \textit{Deletion Through Neutralization}: Negative prompts delete concepts from the generated image through a mutual cancellation effect in latent space with positive prompts. These insights reveal significant potential real-world applications; for example, we demonstrate that negative prompts can facilitate object inpainting with minimal alterations to the background via a simple adaptive algorithm. We believe our findings will offer valuable insights for the community in capitalizing on the potential of negative prompts.
  
\end{abstract}

\vspace{-12pt}
\section{Introduction}
\vspace{-8pt}
  It has been widely acknowledged that diffusion models have made tremendous breakthroughs in image and video generation~\cite{saharia2022photorealistic,rombach2022high,dhariwal2021diffusion,nichol2021glide,ho2022classifier}. Despite their capabilities, these models sometimes produce images that do not fully align with the intended meaning of their textual prompts, motivating a surge in research aimed at enhancing image fidelity and relevance~\cite{chefer2023attend,hertz2022prompt,wang2022diffusiondb}. Notable advancements include the development of classifier-free guidance~\cite{ho2022classifier}, manipulation of cross-attention map~\cite{kawar2023imagic,chefer2023attend}, integration with large language models~\cite{lian2023llm,zhong2023adapter}, and usage of the semantic information of the prompts~\cite{feng2022training,hertz2022prompt}. Among these innovations, the concept of negative prompts --- guiding models by specifying what not to generate --- has gained great attention for its effectiveness~\cite{armandpour2023re,blog2023ryan,blog2023andrew,blog2023kapoor,blog2023woolf}. However, most of the works are merely relying on experimental results and lack a deep understanding on how negative prompts work. Such a lack of analysis of the negative prompts further prevents people from designing more effective negative prompts to obtain better prompts alignment.

\begin{figure}[th]
    \centering    \includegraphics[width=0.95\textwidth]{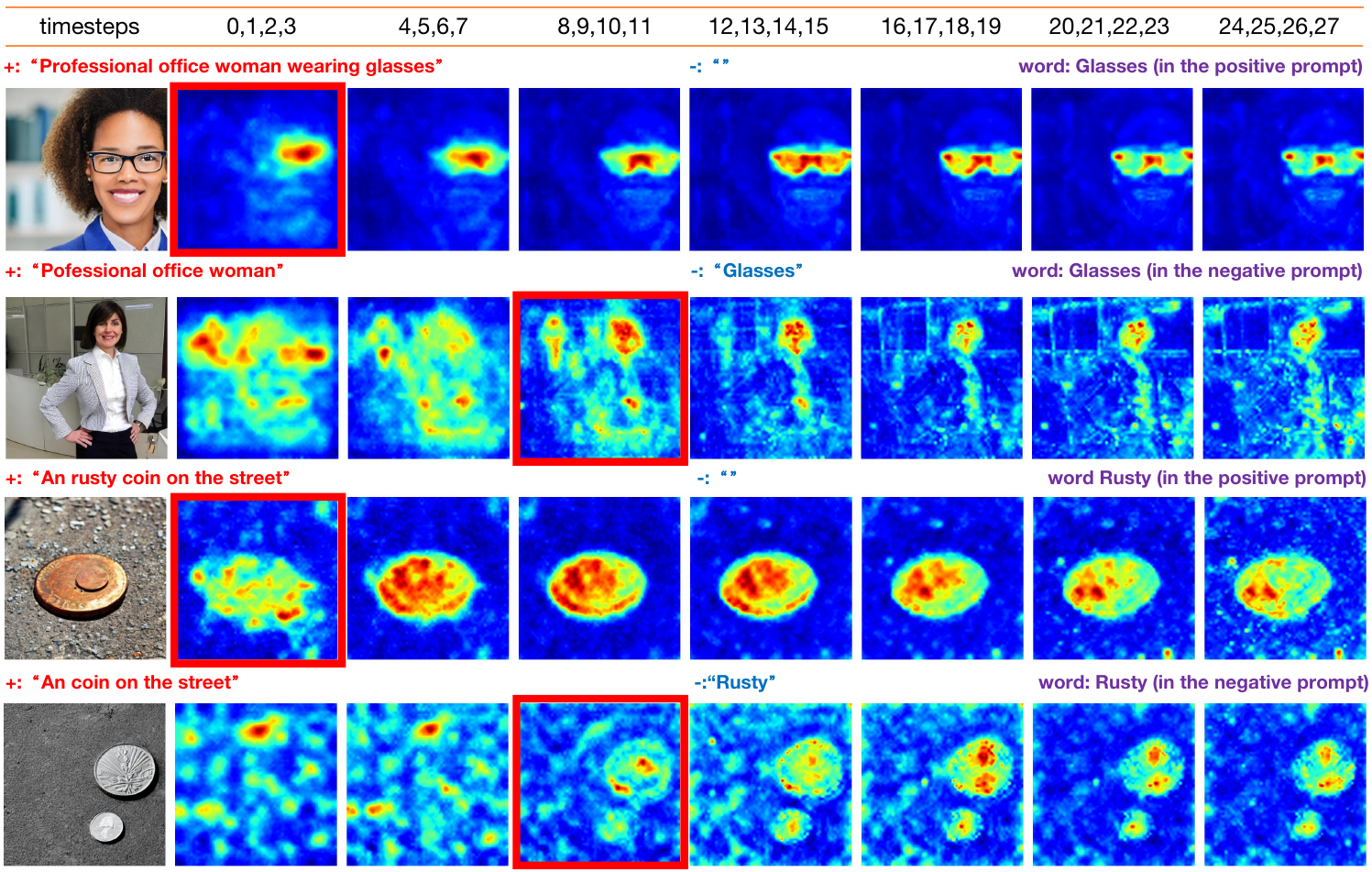} 
    \vspace{-7pt}
    \caption{Illustration on when the negative prompts attend to the "right" place. For example, we consider the face of the person as the "right place"  for the "glasses" token. Every row represents an independent diffusion process where the first and the third rows show the tokens in the \red{positive} prompt and the second and fourth rows visualize those in the \blue{negative} prompt. The \red{positive prompt(+)}, \blue{negative prompt(-)}, and the corresponding \violet{token} of the attention map are listed on top of each of the rows. Every column denotes the different diffusion steps used to visualize the cross-attention heat maps. We also enclose the feature map which attends to the "right" place for the first time, with a square box \boldsquare[red] .}
    \label{fig:when}
    \vspace{-13pt}
  \end{figure}

  In our work, we perform a systemic study on negative prompts to fill this gap. With a focus on the dynamics of the diffusion steps, our central research question is, \textit{"When and how do negative prompts take effect?"}. Our investigation breaks down the mechanism of negative prompts into noun-based removal and adjective-based alteration tasks, leading to intriguing insights through experimentation. Specifically, to investigate when negative prompts start to exert their influence, we analyze the model's cross-attention maps that illustrate the likelihood of specific tokens appearing in the image pixels. We identify the \textbf{critical step} at which negative prompts begin to influence the generation process, highlighting the dramatic difference in how negative and positive prompts operate. The study reveals a significant delay in the critical step of negative prompts compared to positive ones, as clearly illustrated in Figure~\ref{fig:when}. 
  
\begin{figure}[th]
\vspace{-15pt}
    \centering    \includegraphics[width=0.95\textwidth]{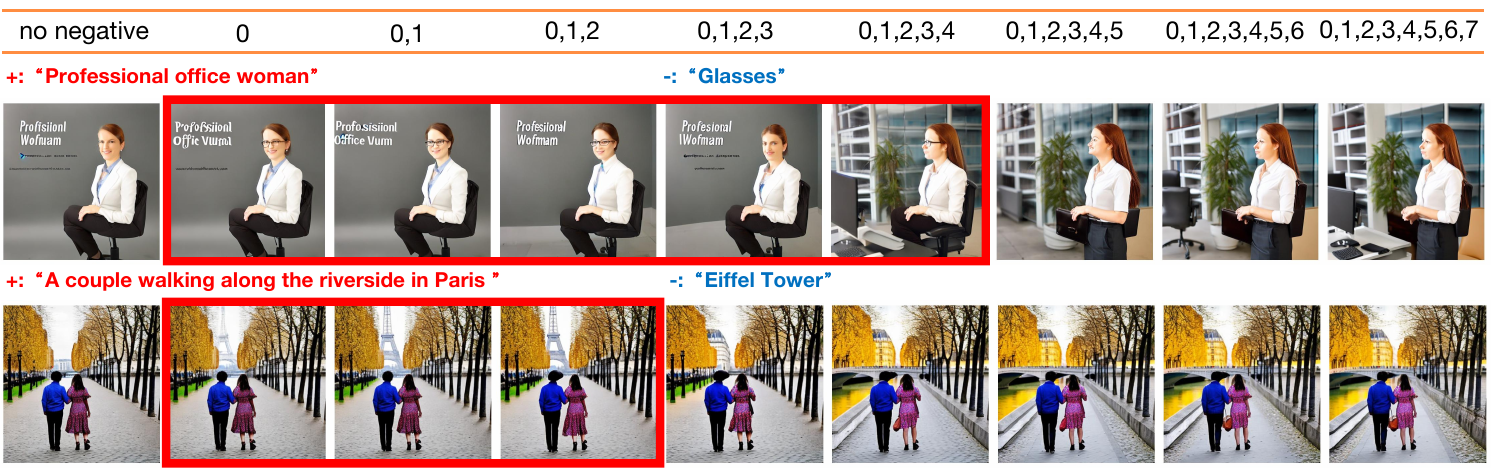} 
    \vspace{-7pt}
    \caption{Illustration: Reverse activation. Each column shows an image generated by applying negative prompts in some specific steps which is shown at the top of the picture. In these two examples, the diffusion process without applying a negative prompt does not produce the object mentioned in the negative prompt. But interestingly, introducing a negative prompt in the early stages results in the generation of the specified object, which is marked with  \boldsquare[red] .}
    \label{fig:how_constra}
    \vspace{-13pt}
\end{figure}

  To figure out how negative prompts take effect, we delve into the architecture of mainstream text-to-image diffusion models to uncover a possible cause: an insufficient exchange of information between the pathways dealing with positive and negative prompts. Analyzing the patterns of estimated noises in object deleting tasks, \textbf{we find that negative prompts initially generate a target object at a specific location within the image, which neutralizes the positive noise through a subtractive process, effectively erasing the object.} Furthermore, we observe a counter-intuitive model behavior ``Reverse Activation'' as shown in Fig~\ref{fig:how_constra}. That is, introducing a negative prompt in the early stages of diffusion paradoxically results in the generation of the specified object in the initial generation stage. We give a detailed explanation for this model behavior based on two findings of diffusion dynamics: the Inducing Effect and the Momentum Effect. The former effect reveals that the negative prompt can induce the positive estimated noises to increase in some specific directions, while the latter shows that the estimated noises tend to keep in the same direction in the diffusion process, which means noises exhibit a significant correlation with their preceding segments. We also point out the potential hazards of applying negative prompts too early, that they may distort the original structure of the image.

  Building on our insights, we introduce a novel controllable inpainting approach aimed at \textbf{deleting undesired elements while preserving as much of the remaining content as possible.} As shown in Fig~\ref{fig:intro}, applying the negative prompts too early can disrupt the layout of an image that has not yet been fully formed. The best timing for introducing these prompts should be after the critical step. Based on the insights, we propose to involve the negative prompts in the middle of the reverse-diffusion process which shows great success. Note that our method does not need any model retraining and modifications in the sampling step during inference.

  Our contributions can be summarized as follows: (1) We have uncovered the critical steps and underlying dynamics that govern the effectiveness of negative prompts (2) We have identified and highlighted the fundamental issue of information lag that occurs between the activation of negative and positive prompts. (3) We provide insights into the strategic design of negative prompts and introduce a novel approach for controllable image inpainting tasks.

\vspace{-8pt}
\section{Related Work}
\vspace{-8pt}
 \textbf{Prompts Analysis:} Since the development of text-to-image diffusion models, there has been a surge of interests in understanding its image generation mechanism through the lens of prompts.
 Tang~\cite{tang2022daam} employed cross-attention maps to analyze prompts through the lens of computational linguistics. Hertz~\cite{hertz2022prompt} revealed that cross-attention layers are imbued with significant semantic content derived from text prompts. Tumanyan~\cite{tumanyan2023plug} demonstrated that self-attention layers encode layout information. Furthermore, Balaji~\cite{balaji2022ediffi} and Mahajan~\cite{mahajan2023prompting} showed that different stages in the process focus on different kinds of features including color, texture, and shape. In contrast to the extensive focus on positive prompts, research on negative prompts remains unexplored, let alone their dynamics along the temporal dimension. Our research primarily delves into negative prompts, exploring the interplay between negative and positive prompts through the lens of temporal evolution.

 \noindent\textbf{Object Removal:} Object removal is the process of eliminating undesired objects from an image. Criminisi~\cite{criminisi2004region} initially conceptualized object removal as an image inpainting task. Yu~\cite{yu2018generative}proposed a novel deep generative model-based approach that can synthesize novel image structures and utilize surrounding image features to boost performance. Yildirim~\cite{yildirim2023inst} trained a diffusion model that can remove objects based on the instructions given as text prompts. Yang~\cite{yang2023magicremover} introduces an attention guidance strategy to constrain the sampling process of diffusion models to enable efficient removal. Existing methods primarily aim to delete a specific object from a given image. In contrast, our approach adopts a different perspective, tailored to the needs of text-to-image model users. Starting with a textual description, we initially create an image using text-to-image diffusion models. Subsequently, we adjust the text to produce a revised image, effectively removing any undesired objects present in the original. There have already been some attempts in this field. For example, Woolf~\cite{blog2023woolf}, O'Connor~\cite{blog2023ryan}, Andrew~\cite{blog2023andrew} and Kapoor~\cite{blog2023kapoor} have proposed to utilize negative prompts to do the task. However, their approaches tend to significantly alter the context of the generated image compared to the original. In contrast, our findings suggest that negative prompts can be used to effectively remove the object, while at the same time preserving the background information if they are applied only within a critical period of the generation process. Moreover, this approach is training-free and assumes zero modification to the structure of diffusion models.
\vspace{-8pt}
 \section{Preliminary}
 \vspace{-8pt}
\textbf{Denoising Diffusion Probabilistic Models (DDPM)} 
As a new family of powerful generative models, diffusion models could achieve superb performance on high-quality image synthesis.
The complete modeling of DDPM consists of \textit{1). a forward process} and \textit{2). a reverse process}.
Given a sample from data distribution $\textbf{x}_0 \sim p_{data}(\textbf{x})$, the forward process gradually injects Gaussian noise to the original data ($\textbf{x}_0$):
\begin{align}
    q(\textbf{x}_t|{\textbf{x}_{t-1}}) = \mathcal{N}\big(\textbf{x}_{t-1};\sqrt{1-\beta_t}\textbf{x}_{t-1},\beta_t\textbf{I}\big), 
\end{align}
where $\beta_t$ is a scheduler designed so that the Markov chain converges to standard Gaussian noise ($\textbf{x}_T \sim \mathcal{N}$) after $T$ steps.
The reverse process then starts with this standard Gaussian noise and repeatedly applies a model ($\theta$) to denoise it back to the real data:
\begin{align}
p_\theta(\textbf{x}_{0:T}) = p(\textbf{x}_T)\prod_{t=1}^Tp_\theta(\textbf{x}_{t-1}|\textbf{x}_t),\\ \text{where } p_\theta(\textbf{x}_{t-1}|\textbf{x}_t) = \mathcal{N}(\textbf{x}_{t-1};\mu_\theta(\textbf{x}_t,t),\Sigma).
\end{align}

\noindent\textbf{Classifer-free guidance for conditional generation}
Text-to-image diffusion models introduce the classifier-free context information into the reverse diffusion process through cross attention map.
At each sampling step, the predicted error is obtained by subtracting the unconditional error from the conditional error with a guidance strength $w$:
\begin{align}  \hat{\epsilon}_\theta((\textbf{x}_t),c(s),t)= (1+w)\epsilon_\theta(\textbf{x}_t,c(s),t) - w\epsilon_\theta(\textbf{x}_t,c(\emptyset),t),
\end{align}
where $c(s)$ is the conditional signal of text $s$, $c(\emptyset)$ is obtained by passing an empty string to the text encoder. 

\noindent\textbf{Negative prompts}\label{sec:notion negative prompts}
Woolf\cite{blog2023woolf} finds that the generative process could be better guided with text prompts that instruct the AI model that it should not include certain elements in its generated images.
Specifically, when the empty string $\emptyset$ in the unconditional error is replaced by an actual prompt, it represents what to remove from the (generated) image due to the negative sign.
This can be formally written as:
\begin{align}  \hat{\epsilon}_\theta((\textbf{x}_t),c(s),t)= (1+w)\epsilon_\theta(\textbf{x}_t,c(p_{+}),t) - w\epsilon_\theta(\textbf{x}_t,c(p_{-}),t),
    \label{eq:negative_prompt}
\end{align}
where $p_{+}$ is the regular user prompt (positive prompt) and $p_{-}$ is the negative prompt.


\noindent\textbf{Stable diffusion}~\cite{rombach2022high} is a latent text-to-image diffusion model featuring processing over a lower dimensional latent space to reduce memory and compute complexity. In our experiments, we adopt Stable Diffusion v2~\cite{Rombach_2022_CVPR} provided by diffusers~\cite{von-platen-etal-2022-diffusers} and set the diffusion steps as 30 in all of the experiments.


\vspace{-8pt}
\section{When do negative prompts take effect}\label{sec:when}
\vspace{-8pt}

\subsection{Qualitive Analysis}
\vspace{-8pt}


\noindent\textbf{Visualising cross-attention maps across diffusion steps.} In conditional diffusion models, cross-attention layers contextualize text embeddings with coordinate-aware latent representations of the image and output scores for each token-image patch pair. Hence, each element in the cross-attention map can be viewed as the probability that the specific token appears in that position. Following the approach of Daam~\cite{tang2022daam}, we gather the scores from various layers for each token we focus on. Then we resize the feature maps to the same size and average them. Notably, different from Daam which averages the scores across all the time steps, we collect and present maps of different steps individually. These heat maps are then organized into sets of four and shown in Fig~\ref{fig:when}.


\noindent\textbf{There exists a delay in the effect of negative prompts following the impact of positive prompts for both nouns and adjectives.}
As shown in Fig~\ref{fig:when}, we observe a delayed effect of negative prompts.
Take the images of "woman with glasses" as an example (top 2 rows).
In the first row, the glasses in the positive prompt are correctly attended to the woman's head from the very beginning, within the first four steps.
Conversely, the glasses in the negative prompt cannot attend to the right position until the eighth step.
Intuitively, this delay stems from classifier-free guidance in Equation~\eqref{eq:negative_prompt}: at every step, both negative and positive prompts attend to the same noise map independently, with their interaction occurring only indirectly after the subtraction step.
As a result, the negative prompt has to wait for the target object (woman's face) specified in the positive prompt to appear before it can attend to it.
The above analysis also applies to the case where the negative prompt specifies an adjective (bottom two roles): The negative prompt "rusty" can only attend to the coin after the coin is been generated.

\subsection{Quantitative Analysis}
\begin{figure}[thbp]
    \centering
    \begin{subfigure}[b]{0.48\textwidth}
\includegraphics[width=\textwidth]{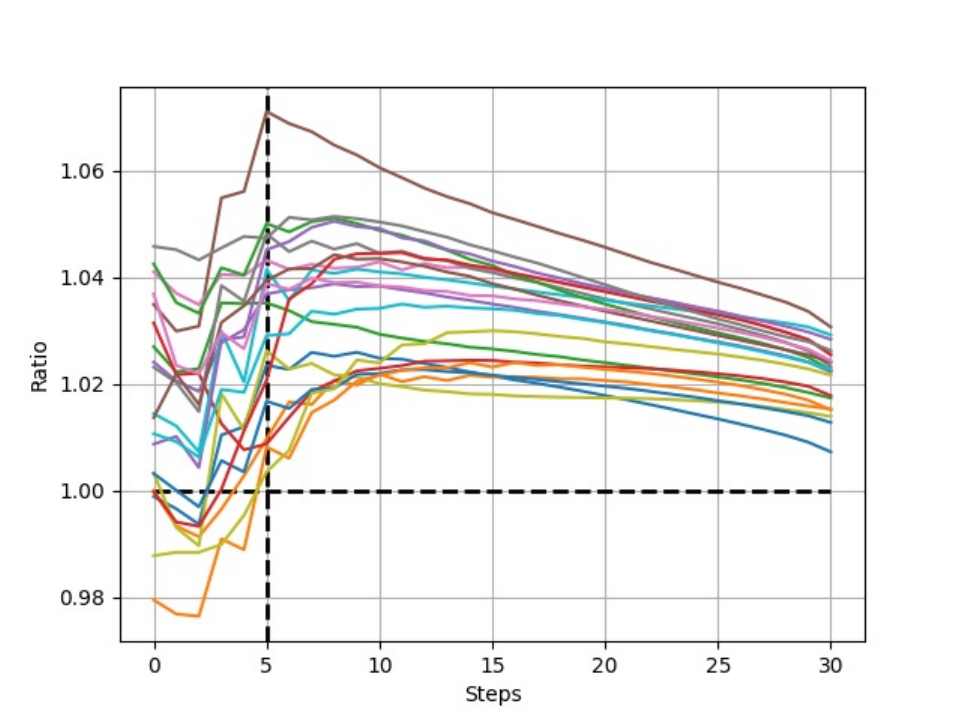}
        \caption{Nouns for removal task}
        \label{fig:when_ratio_noun}
    \end{subfigure}
    \hfill 
    \begin{subfigure}[b]{0.48\textwidth}
        \includegraphics[width=\textwidth]{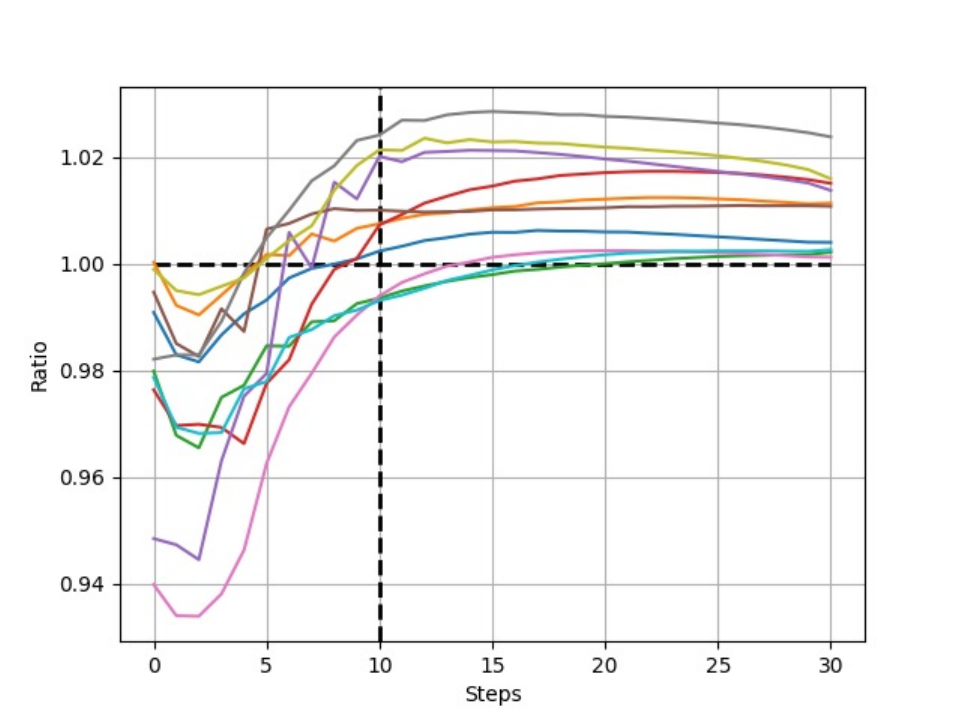}
        \caption{Adjectives for picture refinement}
        \label{fig:when_ratio_adj}
    \end{subfigure}
    \vspace{-7pt}
    \caption{Illustration of Effectiveness of Negative Prompts Over Time. The x-axis represents the time step. The y-axis denotes the strength of the negative prompt. In the left figure, there is a peak at the 5th step for the noun-based negative prompt, indicating the critical step. Meanwhile, In the right figure, we observe a plateau around the 10th, as the object have been generated and the negative prompt begins to take effect.}
    \label{fig:when_ratio}
    \vspace{-8pt}
\end{figure}

\vspace{-8pt}
In this subsection, we quantify the exact step at which the word in the negative prompt aligns accurately with the target objective. \\
\noindent\textbf{New metric to measure the strength of the negative prompt}. Denoting cross attention map of the k-th layer for the i-th token in the text s at time step $t$, as $F_{k,s(i)}^{(t)}$, we further define the strength of negative prompt at step $t$ as the ratio of the mean of the squared value of the heat maps of the negative prompt at step $t$ to that of the positive prompt. Notably, we specifically select the token in the positive prompt that is most relevant to the negative prompt and disregard the unrelated ones. For example, we choose the word "woman" in the positive prompt to compare with the negative prompt "glasses".
The ratio can be formulated as:
\begin{align}
    r_t = \frac{\Sigma_{k}\|F_{k,p_-(i)}^{(t)}\|_F}{\Sigma_{k}\|F_{k,p_+(r(i))}^{(t)}\|_F}
\end{align}
where $p_+$, and $p_-$ denote the positive and negative prompt respectively. And $r(\cdot)$ represents the mapping function from the negative prompt to its most relevant token in the positive prompt. We categorize the negative prompts into two distinct groups for our examination: nouns, such as 'glasses', which are utilized in object removal tasks, and adjectives, like 'ugly', aimed at refining the visual quality of images. we select a set of 10 corresponding prompt pairs. The experiments are conducted across 10 distinct random seeds to ensure robustness in our findings. We plot the $r_t \sim t$ curves averaged on the seeds in Fig~\ref{fig:when_ratio}.
\noindent\textbf{Peak at the 5th step for the noun-based negative prompt} as illustrated in Fig~\ref{fig:when_ratio_noun}, indicating a critical step here. Initially, the ratio is near 1, possibly due to the Unet framework treating negative and positive prompts with parity, as discussed in Section~\ref{sec:notion negative prompts}. At that time, the negative prompt wants to generate some objects in the middle of the pic regardless of the context of the positive prompt. As we approach the peak, the negative prompt begins to assimilate layout cues from its positive counterpart, trying to remove the object. This results in the peak representing the zenith of its influence. Following this, as the element gradually disappears from the image, the impact of the negative prompt diminishes, with no remaining elements in the image to trigger the neural response.
\noindent\textbf{A plateau around the 10th step for adjective-based negative prompts} is depicted in Fig~\ref{fig:when_ratio_adj}, indicating the existence of the critical step. During the initial stages, the absence of the object leads to a subdued response, with the strength below one. Between the 5th and 10th steps, as the object becomes clear, the negative prompt accurately focuses on the intended area and maintains its influence.

\vspace{-8pt}
\section{How do negative prompts take effect}\label{sec:how}
\vspace{-12pt}
To examine the dynamics in the reverse-diffusion process, we focus on analyzing the series of the estimated noises $\{\epsilon^{(t)}\}_{t=0}^T$.

\subsection{Neutralization Hypothesis} 
\begin{figure}[t]
    \centering    \includegraphics[width=0.95\textwidth]{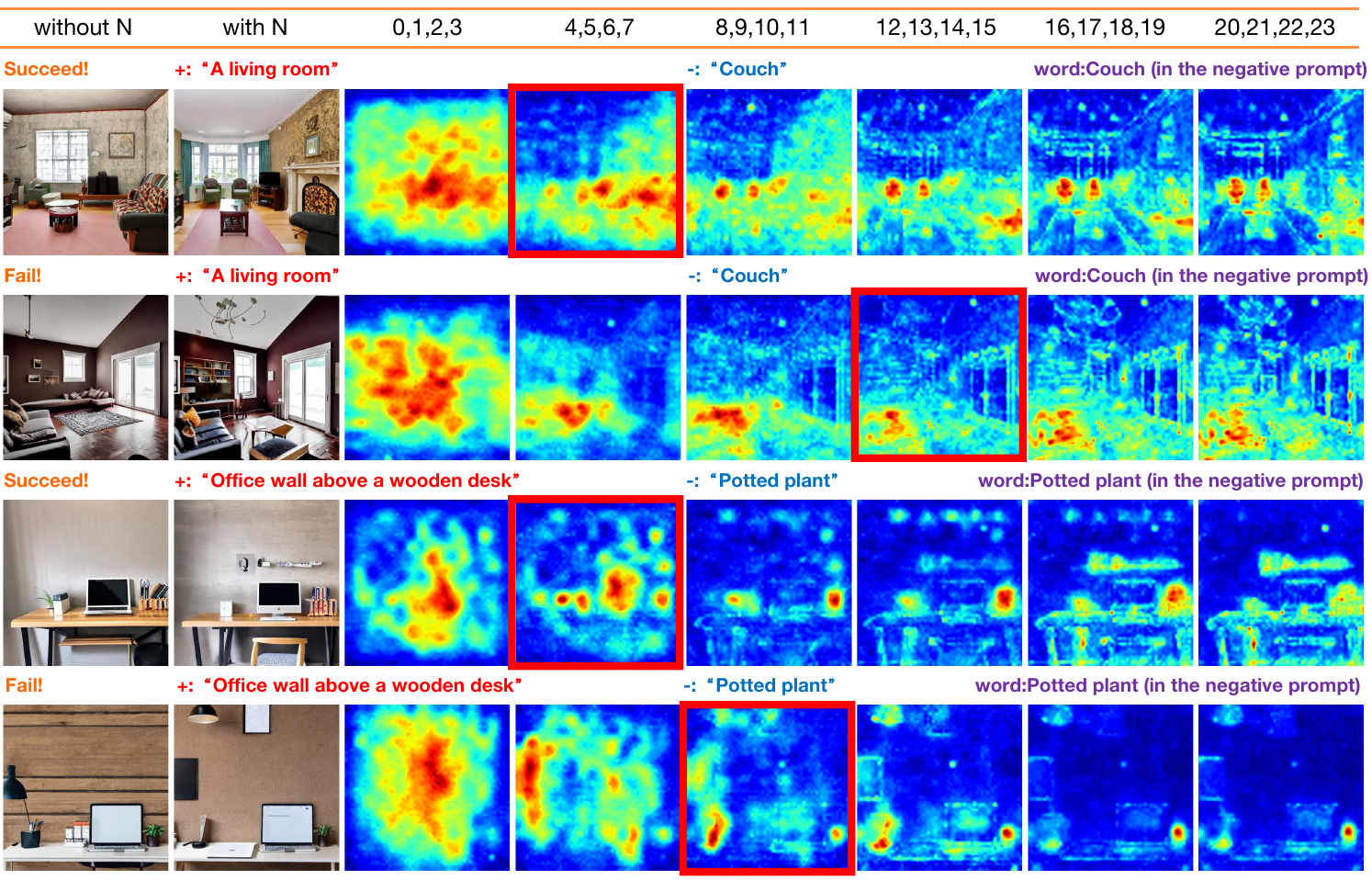} 
    \vspace{-7pt}
    \caption{Illustration: Heat maps showcasing the outcomes of object removal using negative prompts, with both successes and failures. Successful removals are placed in the first and third rows, while the failed attempts occupy the second and fourth rows.  The first column shows the pictures \textbf{without} applying negative prompts contrasted by the second column, which features images \text{with} negative prompts. Notably, the feature map that first targets the relevant location is marked by a red square box \boldsquare[red] . It's evident that the successful cases exhibit earlier attention to the target areas.}
    \label{fig:when_fail}
    \vspace{-10pt}
\end{figure}

We hypothesize that the negative prompts perform deletion through the canceling effect, where positive and negative noises align and nullify each other post-subtraction in Equation~\eqref{eq:negative_prompt}. Supporting this, we observe in the failure cases of object deletion depicted in Fig.~\ref{fig:when_fail}. A successful case is in the third row, where the attention map starts to target the location of a potted plant between steps 4-7, effectively counteracting the positive noise that would otherwise materialize a potted plant. Conversely, in the bottom row, attention doesn't focus on the lower right corner until step 8. By this stage, with the object nearly fully formed, it's too late for effective cancellation—resulting in only minimal impact on the finer details.

\vspace{-8pt}
\subsection{Reverse Activation} 
\vspace{-8pt}
 The phenomenon of Reverse Activation is observed when a negative prompt, introduced in the early stages of the diffusion process, unexpectedly leads to the generation of the specified object within the context of that negative prompt.  In contrast, omitting negative prompts results in the absence of the object. As demonstrated in Figure~\ref{fig:how_constra}, if we apply "Glasses" as negative prompts in the first 3 steps, it will generate the glasses in the final output. In this section, our goal is to shed light on this phenomenon by analyzing the mechanism behind negative prompts. We start by examining the data distribution, highlight two intriguing observations, and ultimately offer an explanation.
 
\begin{wrapfigure}{h}{0.57\textwidth}
\centering
\includegraphics[width=0.95\linewidth]{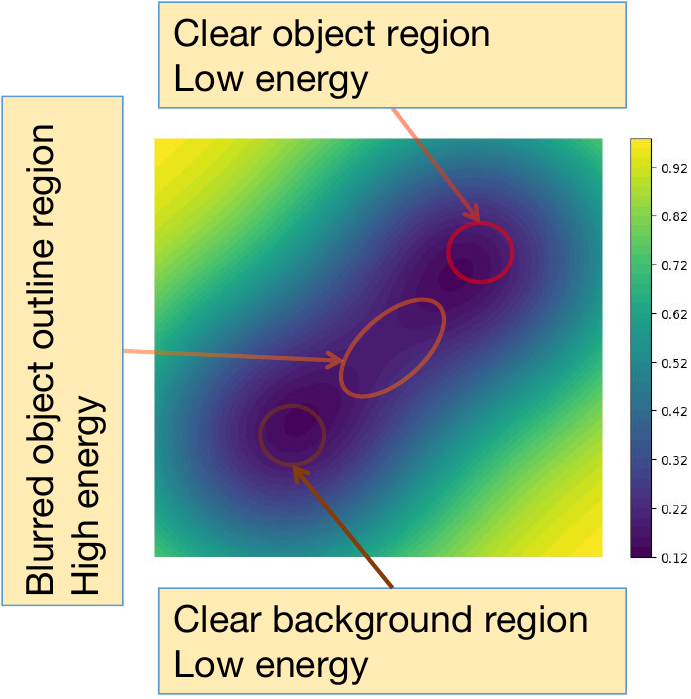}
\caption{Illustration: The energy function in the image generation dynamics. The value at the pixel represents the energy of the point in the data distribution. We mark the \darkbrown{background region}, \red{clear object region}, and \orange{blurred object outline region} by circles. To generate an object from the background, the model should overcome the energy barrier of the \darkbrown{blurred object outline region}. }
\vspace{-8pt}
\label{fig:how_dis}
\end{wrapfigure}
\begin{figure}[th]
    \centering
    \includegraphics[width=0.95\textwidth]{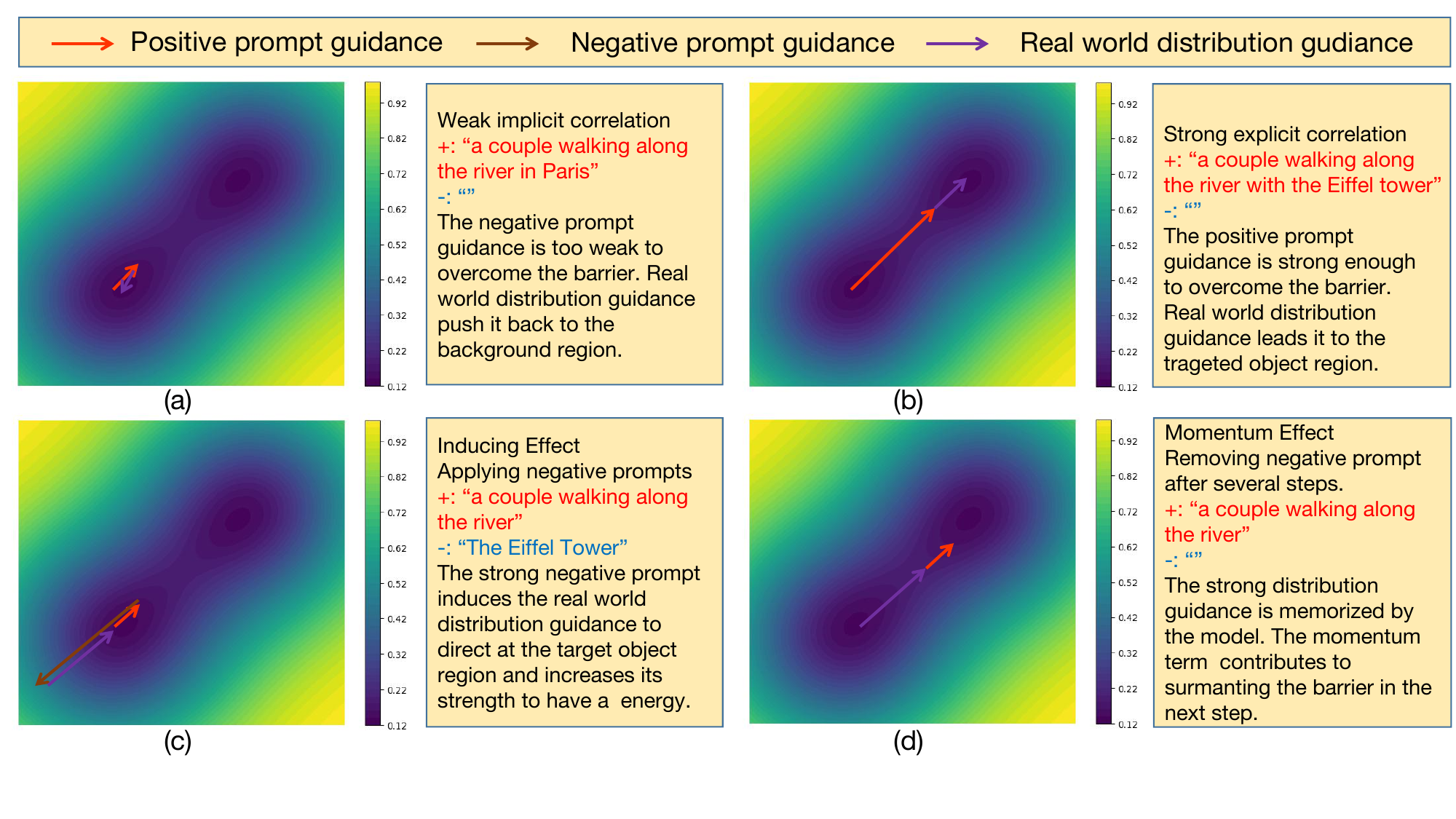} 
    \caption{Illustration: Distinct kinds of guidance. The \violet{purple} arrow shows the guidance of the data distribution, which is the slope of the energy function. The \red{red}, \darkbrown{brown} arrow shows the guidance of the \red{positive} and \darkbrown{negative} prompts respectively.}
    \label{fig:how_four}
    \vspace{-13pt}
\end{figure}

\noindent\textbf{Guidance signals} We borrow the concept of the energy function from Energy-Based Models, as shown in Fig~\ref{fig:how_dis} to represent the data distribution. The function is designed to assign lower energy levels to more 'likely' or 'natural' images according to the model's training data, and higher energy levels to less likely ones. As Real-world distributions often feature elements like a clear blue sky or other uniform backgrounds, alongside distinct objects such as the Eiffel Tower, these elements typically possess low energy scores, making the model inclined to generate them. To synthesize a specific object like a tower from scratch, the diffusion process necessarily traverses through an intermediary phase that represents a blurry outline of the object.
Given that such blurry representations are atypical in the training data, they present an 'energy barrier' that hinders the seamless generation of the desired object. So the model requires the guidance of prompts to surmount this barrier. We delve into the dynamics of distinct types of guidance as depicted in Figure~\ref{fig:how_four}. To begin, Figure~\ref{fig:how_four}\red{a} demonstrates that in the absence of explicit guidance, the model struggles to overcome the energy function's barrier, influenced by the natural data distribution, leading it back to generic backgrounds. Conversely, as depicted in Figure~\ref{fig:how_four}\red{b}, when explicit guidance is provided through the inclusion of the object in the positive prompt's context, the model manages to surpass the barrier, with real-world distribution guidance steering it towards the object region. 

\noindent\textbf{Inducing Effect} The intriguing part is observed in Fig~\ref{fig:how_four}\red{c}. Here, we illustrate an instance where direct negative prompts are applied, yet the context is absent from the positive prompt. As a result, the negative prompt guidance is much stronger than the positive prompt guidance, making this point at a considerable distance in the region opposite to the object. \textbf{Consequently, the distribution guidance demonstrates a substantial alignment towards the object and its surrounding area.} Without this, a tower-like structure would emerge at that location. This is because adding or subtracting tower-like features against a uniform backdrop equally contributes to the formation of a tower pattern. As the distribution guidance is encoded into the estimated noises, projecting the positive noise towards the background-to-object direction reveals an enhanced effect in this direction, as opposed to the scenario without a negative prompt in  ig~\ref{fig:how_four}\red{a}. We term the phenomenon as the "Inducing Effect", indicating that the negative prompt triggers the positive noise in a direction that represents the context of the negative prompt.

\noindent\textbf{Simulation Experiment} To validate our hypothesis on the inducing effect, we conducted a follow-up quantitative experiment employing a variety of prompt pairs. Initially, we generate an image using positive prompt $p_+$ along with an empty unconditional prompt $p_e$ and record the series of positive noises $\{\epsilon_{p_+}^{(t)}(p_+,p_e)\}_{t=1}^T$. Simultaneously, we calculate a series of negative noises, $\{\epsilon_{p_-}^{(t)}(p_+,p_e)\}_{t=1}^T$ but refrain from applying it during the sampling process. Following this, we generate additional noise series, $\{\epsilon_{p_+}^{(t)}(p_+,p_-)\}_{t=1}^T$ and $\{\epsilon_{p_+}^{(t)}(p_+,p_-)\}_{t=1}^T$ by applying both the positive prompt $p_+$ and the negative prompt $p_-$ this time. To verify the existence of induction, we project the positive noise onto the negative noise which represents the direction towards the object region, and compute the difference between the two sets. The computation can be formulated as:
\begin{align}
    P_{Ind}^{(t)} &= \frac{<\epsilon_{p_+}^{(t)}(p_+,p_-),\epsilon_{p_-}^{(t)}(p_+,p_-)>}{||\epsilon_{p_-}^{(t)}(p_+,p_-)||^2}\epsilon_{p_-}^{(t)}(p_+,p_-) \\
    P_{Ori}^{(t)} &= \frac{<\epsilon_{p_+}^{(t)}(p_+,p_e),\epsilon_{p_-}^{(t)}(p_-,p_e)>}{||\epsilon_{p_-}^{(t)}(p_+,p_e)||^2}\epsilon_{p_-}^{(t)}(p_+,p_e) \\
    D^{(t)} &= P_{Ind}^{(t)} - P_{Ori}^{(t)}
\end{align}

\begin{figure}[h]
  \centering
  \begin{subfigure}[b]{0.48\textwidth}
    \includegraphics[width=\textwidth]{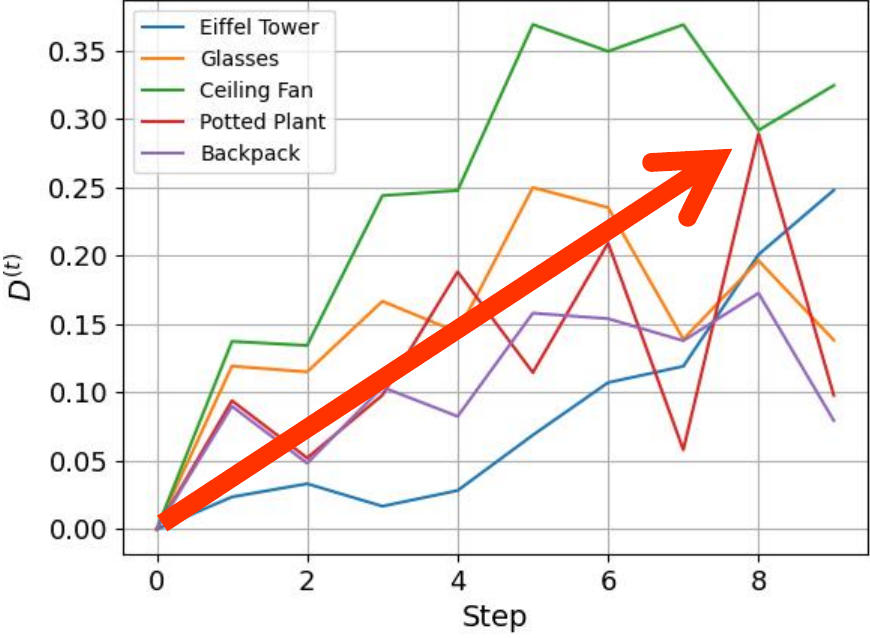}
    \caption{Inducing Effect. The x-axis shows the time step while the y-axis quantifies the amount in the direction towards the object. The upward trajectory of the \textbf{\red{red arrow}} verifies the Inducing Effect phenomenon.}
    \label{fig:how_induce}
  \end{subfigure}
  \hfill
  \begin{subfigure}[b]{0.48\textwidth}
    \includegraphics[width=\textwidth]{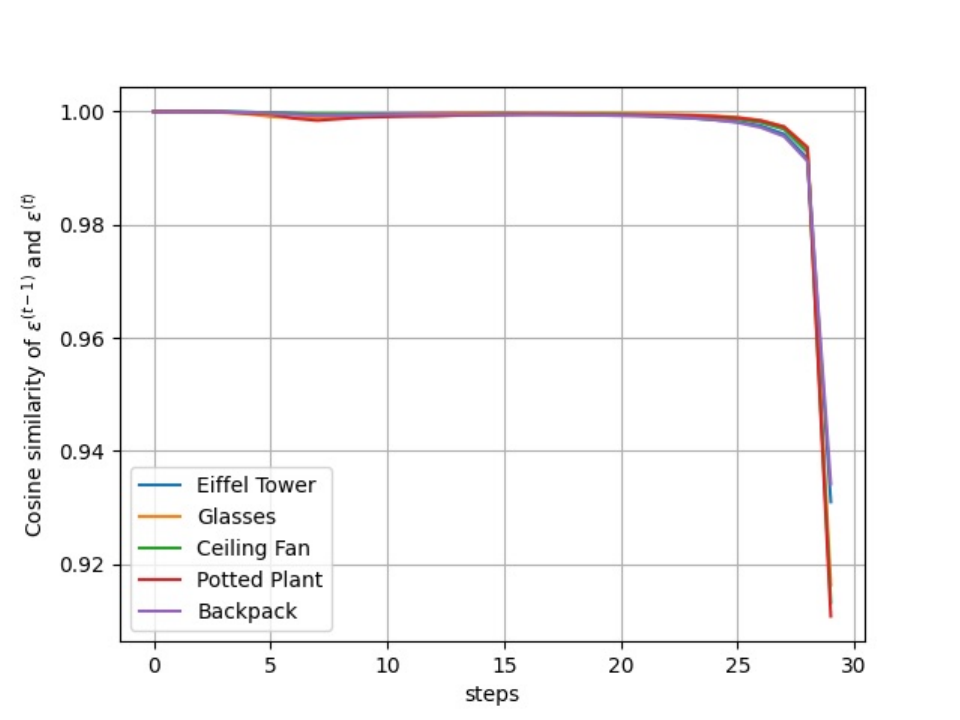}
    \vspace{-5pt}
    \caption{Momentum Effect. The x-axis denotes the time step. The y-axis measures the cosine similarity between noises at consecutive steps. The diffusion process reveals a strikingly high correlation up to 99.5\% in the first 25 steps.}
    \label{fig:how_align}
  \end{subfigure}
  \caption{Illustration: Inducing Effect and Momentum Effect}
  \vspace{-10pt}
  \label{fig:test}
\end{figure}

Intuitively, $P_{Ind}^{(t)}$ and $P_{Ori}^{(t)}$ shows the \violet{distribution guidance} in Fig~\ref{fig:how_four}\red{a} and Fig~\ref{fig:how_four}\red{c}, respectively. And $D^{(t)}$ suggests the difference in the objection direction if a negative prompt containing the object is applied. \textbf{Therefore, a positive difference implies that the presence of the negative prompt effectively induces the inclusion of this component in the positive noise.} We perform experiments on 5 prompt pairs. We run 10 random seeds for each pair, average the results, and plot the $D^{(t)} \sim t$ curve in Fig~\ref{fig:how_induce}. The results demonstrate that the presence of a negative prompt promotes the formation of the object within the positive noise, thereby confirming our hypothesis.

\noindent\textbf{Momentum Effect} Additionally, we observe a behavior akin to momentum, where the generated noises appear to sustain their trajectory in a specific direction. To confirm this phenomenon, we calculate the cosine similarity between the noise at step t ${\epsilon}_\theta((\textbf{x}_t),c(s),t)$ and the noise at the preceding step t-1 ${\epsilon}_\theta((\textbf{x}_t-1),c(s),t-1)$. As shown in Fig~\ref{fig:how_align}, there is a notable correlation between each noise and its immediate predecessor, indicating a momentum-like effect.

\noindent\textbf{Explanation} Finally, we can come to an explanation of Reverse Activation in Fig~\ref{fig:how_constra}. Figure~\ref{fig:how_four}\red{a} demonstrates that without the negative prompt, the implicit guidance is insufficient to generate the intended object, explaining why the object fails to appear in the first column of Figure~\ref{fig:how_constra}. Conversely, as illustrated in Figure~\ref{fig:how_four}\red{c}, the application of a negative prompt intensifies the distribution guidance towards the object, which prevents the object from materializing, clarifying the absence of the object in the last several columns of Figure~\ref{fig:how_constra}. Intriguingly, as supported by the Momentum Effect, if we remove the negative prompt after several steps, the real-world distribution guidance will maintain a large component towards the object's direction in the following steps as shown in Fig~\ref{fig:how_four}\red{d}. Such a momentum effect finally facilitates the object's emergence, as shown in the middle columns in Fig~\ref{fig:how_constra}.

\vspace{-12pt}
\section{Enhanced Controllable Inpainting}
\vspace{-8pt}

\begin{figure}[t]
    \centering
    \includegraphics[width=0.95\textwidth]{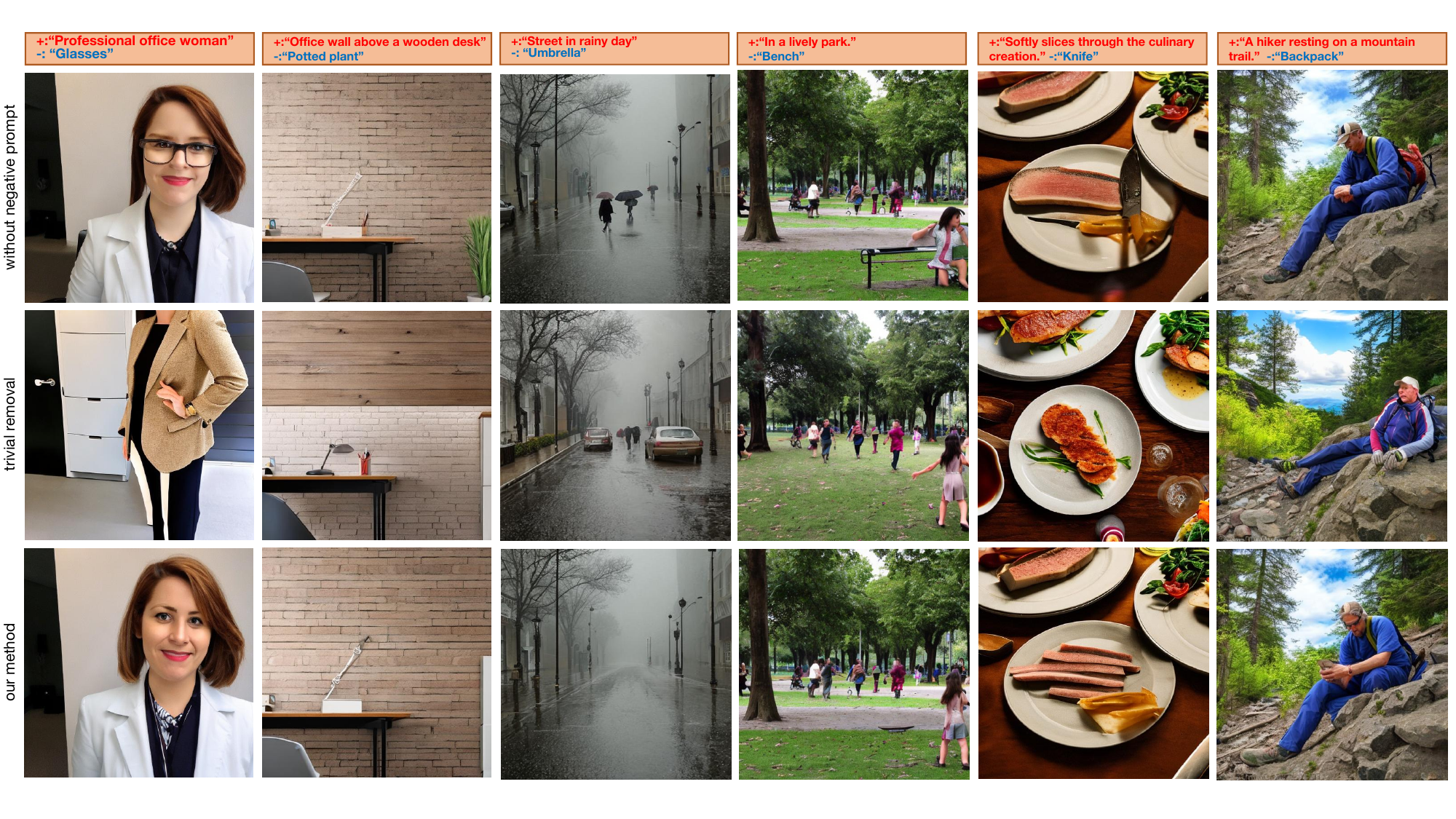}
    \vspace{-15pt}
    \caption{Illustration of our method's ability to remove unwanted objects in the generated image while preserving the main concept, for various combinations of \red{positive(+)} and \blue{negative(-)} prompts. \textbf{From top to bottom}: Initially with solely a positive prompt, followed by the integration of negative prompts throughout all diffusion stages, and finally, applying negative prompts during pivotal stages. Identical seeds were utilized for each column to ensure consistency.}
    \label{fig:intro}
    \vspace{-15pt}
\end{figure}

In this section, we introduce a novel technique for controllable image inpainting that aims to eliminate undesired objects from generated images while preserving the integrity of the original background. Although Woolf~\cite{blog2023woolf} highlights the effectiveness of using negative prompts to remove undesired elements from images, it often leads to substantial modifications to the background, as shown in the second row of Fig~\ref{fig:intro}. 

In Sec~\ref{sec:how}, we claim that the negative prompt takes effect by a neutralization effect. But in Sec~\ref{sec:when}, we observe a notable delay in the activation of negative prompts compared to their positive counterparts. As a result the negative prompts usually don't attend to the right place until step 5, well after the application of positive prompts. Additionally, as depicted in Fig~\ref{fig:how_four}\red{c}, the use of negative prompts in the initial steps can significantly skew the diffusion process, potentially altering the background. This early application throughout the inference process, as practiced by Woolf, could be the reason behind their method's shortcomings. 

To mitigate these issues, we propose to deploy negative prompts post-'critical step' rather than getting it from the beginning. According to our findings, since negative prompts usually don't attend to the interested region until the critical step, all the neutralization and removal would then happen after the critical step. Meanwhile, the later added negative prompt would focus more on the target area with a reduced effect on surrounding regions in the later phase.

\vspace{-8pt}
\subsection{Experiments}
\vspace{-8pt}
In this section, we conduct large-scale experiments to validate the efficacy of our proposed method. \\
\vspace{-12pt}

\begin{wrapfigure}{h}{0.40\textwidth}
\centering
\vspace{-12pt}
\includegraphics[width=0.99\linewidth]{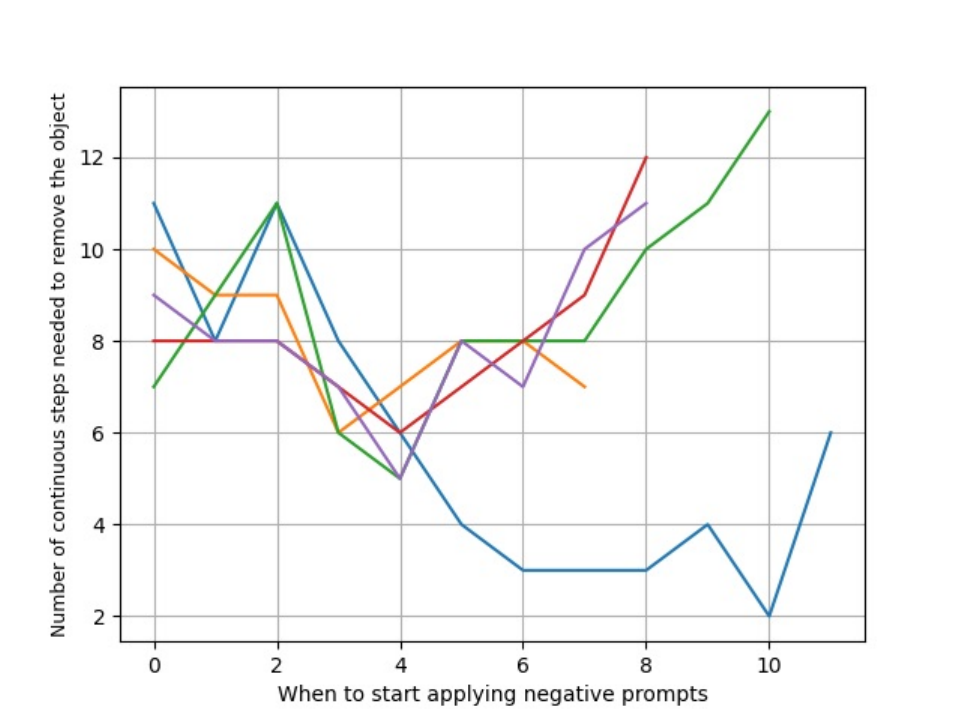}
    \caption{Experiments on finding the best time to apply negative prompts. Each endpoint on the polyline graph represents a specific scenario: the x-coordinate denotes the starting step for applying negative prompts, while the y-coordinate indicates the number of consecutive steps required to precisely remove the targeted object. When the object cannot be removed, the curve terminates.}
\label{fig:apply_step}
\vspace{-26pt}
\end{wrapfigure}

\noindent\textbf{Finding the Timing for Negative Prompts in Inpainting} We tested various combinations of prompts and negative prompts. The results, depicted in Figure~\ref{fig:apply_step}, reveal a U-shaped trend, indicating that employing negative prompts during intermediate steps is most effective. Take the blue line as an example, initiating with negative prompts at the first step necessitates approximately 10 steps for task completion. In contrast, starting at the sixth step significantly reduces this requirement to about 4 steps. Notably, the curves' nadir is around step 5, aligning with earlier insights about the critical step discussed in Section~\ref{sec:when}. Beyond step 11, applying negative prompts appears ineffective in eliminating the desired object. This may be because, in the later stage of diffusion, the shape and structure of the image have been essentially determined. We set steps 5-15 to use negative prompts as default in our later experiments.

\noindent\textbf{Constructing datasets for evaluation of negative prompts} 
Due to the absence of pre-existing datasets in our settings, we adapt available datasets for our needs. We begin by selecting text samples from the COCO~\cite{lin2014coco}, CC~\cite{ng2020cc}, Nocaps~\cite{agrawal2019nocaps}, Places~\cite{zhou2017places}, MSVD~\cite{venugopalan2015msvd}, and Vatex~\cite{wang2019vatex} datasets to serve as prompts for image generation. These images are then analyzed by GPT-4V~\cite{achiam2023gpt} to identify the contained objects. Then we try to remove these objects when generating the image again. For each dataset, we use 1000 text prompts and every prompt is run with 5 different seeds.

\noindent\textbf{Evaluation} We leverage GPT-4V to assess the success of the inpainting process and to determine the relative distance between the original and inpainted images. We also conduct \textbf{human evaluation} for further verification. More details on the evaluation protocols can be found in the Appendix.

\noindent\textbf{Metrics} To assess the efficacy of our approach, we employ two key metrics: The Removal Success Rate(RSR) indicates the success rate of the target object removal.The Relative Removal Success Rate (RRSR) gauges the efficiency of our method relative to a baseline by calculating the ratio of our RSR to the baseline's RSR. A higher RRSR suggests that our method remains effective even when negative prompts are applied over fewer steps. Additionally, the Comparison Rate(CR) measures the extent to which our generated images resemble the original images, as judged by GPT-4V or human evaluators. We first ask GPT4V and humans if images generated by our method are more similar to the origin of one. Then we compute the ratio of affirmative responses to the total number of evaluations. The higher the RR, the better. Details can be found in the Appendix.

\noindent\textbf{Results} Table~\ref{tab:gpt} summarizes the results. As we can see, our method incurs minimal impact on the removal success rate. In fewer than 20\% of instances, our method fails to remove the target object where the baseline method succeeds, addressing concerns that applying negative prompts in fewer steps might compromise inpainting effectiveness. Moreover, on average, our method achieves up to 82.64\% similarity to the original images, underscoring its efficiency. Meanwhile, the results of the human evaluation can be seen in Table~\ref{tab:human} The results show the effectiveness of our method.


\begin{table}[t]
\hfill
	\begin{subtable}[h]{0.47\textwidth}
		\centering
\begin{tabular}{c|c|c|c|c}
    \hline\hline
Dataset & RSR\% & RSR\% & RRSR\% & CR\% \\
\hline
COCO  & 54.41 & 65.02 & 83.67 & 83.32 \\
CC & 54.81 & 64.24 & 85.31& 82.75 \\
MSVD & 55.53 & 66.59 & 83.39 & 83.07 \\
Places & 49.83 & 57.21 & 87.20 & 84.12 \\
Vatex & 61.37 & 71.00 & 86.44 & 82.52 \\
Nocaps & 48.88 & 56.72 & 86.17 & 81.26 \\
\hline
Avg & 54.16 & 63.46 & 85.37 & 82.84 \\
\hline
\end{tabular}
		\caption{Main Results of GPT-4V Evaluation}
		\label{tab:gpt}
	\end{subtable}
 \hfill
	\begin{subtable}[h]{0.47\textwidth}
		\centering
\begin{tabular}{c|c|c|c|c}
    \hline\hline
Dataset & RSR\% & RSR\% & RRSR\% & CR\% \\
\hline
COCO  & 87.27 & 90.90 & 96.01 & 92.68 \\
CC & 59.32 & 71.18 & 83.34 &  97.67 \\
MSVD & 61.11 & 81.48 & 75.00 & 83.67 \\
Places & 53.12 & 76.56 & 69.38 &  90.66 \\
Vatex & 57.57 & 73.53 & 78.29 & 80.12 \\
Nocaps & 72.32 & 88.46 & 81.75 & 87.67 \\
\hline
Avg & 65.11 & 80.35 & 80.62 & 88.45 \\
\hline
\end{tabular}
		\caption{Main Results of Human Evaluation}
		\label{tab:human}
	\end{subtable}
 \hfill
	\caption{For both RRSR and CR, the higher the better. The first column shows the RSR of the baseline by applying negative prompts to all steps. The second column shows the RSR of our method.}
\end{table}
\vspace{-8pt}
\section{Future work}
\vspace{-8pt}
  In our experiments, we focus primarily on tasks involving the removal of nouns and the attribution based on adjectives, deferring the exploration of other parts of speech and tasks to future research. Our findings highlight the challenge of information lag between pairs of positive and negative prompts. A straightforward remedy could involve increasing interactions during the noise generation phase. Additionally, our method of controllable object removal in image generation presents a novel approach for creating image inpainting datasets. Finally, applying negative prompts in the training process as a form of data augmentation may potentially enhance performance further, which is left as future work.
  \vspace{-8pt}
\section{Conclusion}
\vspace{-8pt}
  In conclusion, our research provides a comprehensive analysis of negative prompts in diffusion models for image generation. Through systematic experiments, we have identified the critical steps where negative prompts begin to influence the image generation process, uncovering a significant lag in the transition from positive to negative prompt effects. This insight led us to develop a novel approach that strategically applies negative prompts at an optimal stage in the reverse-diffusion process, ensuring the removal of undesired elements while preserving the image's integrity. Our contributions not only shed light on the underlying dynamics of negative prompts but also offer a practical solution for controllable image inpainting tasks, significantly improving upon existing methods without the need for network retraining or modifications during inference.

\bibliographystyle{plainnat}
\bibliography{main.bib}

\end{document}